\documentclass[runningheads]{llncs}
\usepackage[T1]{fontenc}
\usepackage{graphicx}
\usepackage{amsmath}
\usepackage{booktabs}
\usepackage{algorithm}
\usepackage{algorithmic}
\usepackage{hyperref}
\usepackage{enumitem}
\usepackage{amsfonts}
\usepackage{mathabx}
\usepackage[misc]{ifsym}
\usepackage{color}

\begin{document}
%
\title{WiNet: Wavelet-based Incremental Learning \\
for Efficient Medical Image Registration}
\titlerunning{Wavelet-based Incremental Learning for Efficient Medical Image Registration}
%
\author{Xinxing Cheng\inst{1} \and
Xi Jia\inst{1} \and
Wenqi Lu\inst{2}  \and
Qiufu Li \inst{3} \and
Linlin Shen \inst{3} \and \\
Alexander Krull \inst{1}  \and
Jinming Duan\inst{1}$^{\textrm{(\Letter})}$
}


%
\authorrunning{X. Cheng et al.}
%
\institute{School of Computer Science, University of Birmingham, Birmingham, B15 2TT, UK \\\email{j.duan@cs.bham.ac.uk}\\
\and  Department of Computing and Mathematics,\\ Manchester Metropolitan University, Manchester, M15 6BH, UK
\and  National Engineering Laboratory for Big Data System Computing Technology,\\ Shenzhen University, 518060, China}
%
%
%
\maketitle              
\begin{abstract}
Deep image registration has demonstrated exceptional accuracy and fast inference. Recent advances have adopted either multiple cascades or pyramid architectures to estimate dense deformation fields in a coarse-to-fine manner. However, due to the cascaded nature and repeated composition/warping operations on feature maps, these methods negatively increase memory usage during training and testing. Moreover, such approaches lack explicit constraints on the learning process of small deformations at different scales, thus lacking explainability. In this study, we introduce a model-driven WiNet that incrementally estimates scale-wise wavelet coefficients for the displacement/velocity field across various scales, utilizing the wavelet coefficients derived from the original input image pair. By exploiting the properties of the wavelet transform, these estimated coefficients facilitate the seamless reconstruction of a full-resolution displacement/velocity field via our devised inverse discrete wavelet transform (IDWT) layer. This approach avoids the complexities of cascading networks or composition operations, making our WiNet an explainable and efficient competitor with other coarse-to-fine methods. Extensive experimental results from two 3D datasets show that our WiNet is accurate and GPU efficient.  Code is available at \url{https://github.com/x-xc/WiNet}.

\end{abstract}
\section{Introduction}

Deformable image registration is an essential step in many medical imaging applications \cite{sotiras2013deformable,yang2014automated}. Given a moving and fixed image pair, deformable registration estimates a dense non-linear deformation field that aligns the corresponding anatomical structures. Conventional methods such as LDDMM \cite{Beg2005}, DARTEL \cite{DARTEL}, SyN \cite{pmid17659998}, Demons \cite{VERCAUTEREN2009S61} and ADMM \cite{thorley2021nesterov} are time-consuming and computationally expensive, due to instance-level (pair-wise)  iterative optimization. Nevertheless, such methods may involve sophisticated hyperparameter tuning, limiting their applications in large-scale volumetric registration.



U-Net-based methods have recently dominated medical image registration due to their fast inference speed. Following the generalized framework of VoxelMorph \cite{balakrishnan2019voxelmorph}, some methods have been proposed by adding stronger constraints over the deformation field such as inverse/cycle consistency \cite{zhang2018inverse,kim2021cyclemorph} and diffeomorphisms \cite{dalca2018unsupervised,dalca2019unsupervised}. Another track of methods proposed more advanced neural blocks such as vision-transformer \cite{chen2021vit,chen2022transmorph}  to model long-range information but inevitably increases the computational cost, sacrificing the training and testing efficiency. To reduce the repeated convolution operations in U-Net architecture and fasten the speed, model-driven methods such as B-Spline \cite{qiu2021learning}  and Fourier-Net \cite{jia2023fourier} have proposed to learn a low-dimensional representation of the displacement. However, Fourier-Net only learns the low-frequency components of the displacement and B-Spline interpolates the displacement from a set of regular control points, therefore deformation fields estimated by such methods lack local details, limiting its applications to large and complex registrations.



Recent works \cite{de2019deep,zhao2019unsupervised,zhao2019recursive,mok2020large,modelT,hu2022recursive,kang2022dual} progressively estimate the large and complex deformations using either multiple cascades (where each cascade estimates a small decomposition of the final deformation) or pyramid \textbf{coarse-to-fine} compositions of multi-scale displacements within one network. Though the cascaded methods \cite{de2019deep,zhao2019unsupervised,zhao2019recursive} show improved registration accuracy for estimating large deformation by composing small deformations, their computational costs increase exponentially with more cascades. Pyramid methods \cite{mok2020large,kang2022dual,hu2022recursive,modelT}, on the other hand, estimate multi-scale deformations and sequentially compose them to the final deformation. For instance, LapIRN \cite{mok2020large} utilizes a Laplacian pyramid network to capture large deformations by composing three different scale flows. However, the optimal performance for LapIRN requires sophisticated iterative training of different scales, negatively affecting its training efficiency. PRNet++ \cite{kang2022dual} employs a dual-stream pyramid network for coarse-to-fine registration through sequential warping on multi-scale feature maps, while the adaptation of local 3D correlation layers massively increases memory usage and computational costs. ModeT \cite{modelT} has introduced a motion decomposition transformer that utilizes neighborhood attention mechanisms to first estimate multi-head multi-scale deformations from two-stream hierarchical feature maps. It then employs weighting modules to fuse multi-head flows in each scale and generates the final deformation by composing the fused output from all scales.


In this work, we propose a model-driven multi-scale registration network by embedding the discrete Wavelet transform (DWT) as prior knowledge, which we term WiNet. Compared to model-driven B-Spline \cite{qiu2021learning} and Fourier-Net \cite{jia2023fourier}, our WiNet learns a series of DWT  coefficients that inherently preserve high-frequency local details. Unlike pyramid methods \cite{mok2020large,kang2022dual,hu2022recursive,modelT}, which rely solely on optimizing unsupervised training loss to estimate final deformation without explicitly constraining the learning process of small deformations at each scale, our proposed WiNet incrementally learns scale-wise DWT coefficients and naturally forms full-resolution deformation with a model-driven IDWT layer. Moreover, existing pyramid methods \cite{mok2020large,kang2022dual,hu2022recursive,modelT} require significant memory usage as they need to co-register multi-scale high-dimensional features, preventing their application on GPUs with less memory. 
In contrast, our network avoids compositions on feature maps and estimates only low-resolution DWT coefficients from low-resolution frequency components, making it memory-efficient.

\begin{figure*}[t!]
    \centering
    \includegraphics[width=1\textwidth]{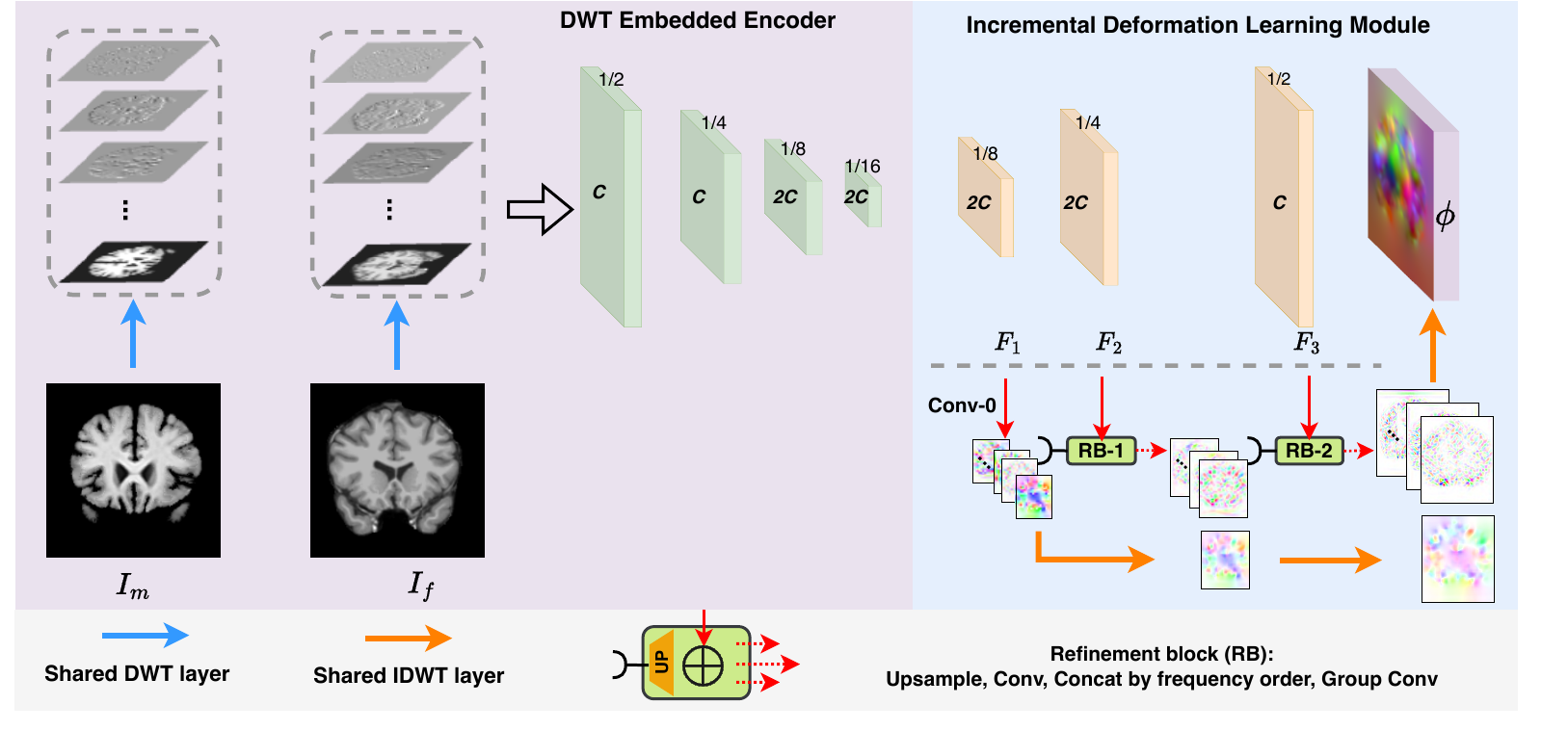}
    \caption{ Architecture of WiNet. Its encoder includes a shared DWT layer and four convolutional contracting layers. The decoder consists of three convolutional expansion layers and an Incremental Deformation Learning Module (which converts the convolutional features into final displacement  $\boldsymbol{\phi}$ with a conv-layer (Conv-0), two refinement blocks (RB-1 and RB-2), and a shared parameter-free IDWT layer). Note that the 3D images, features, and displacements are shown in 2D for illustration. }

    
    \label{fig:model}
\end{figure*}

WiNet contains a convolutional encoder-decoder network and an incremental deformation learning module designed to explicitly learn distinct frequency coefficients at various scales. Specifically, the initial decoding layer simultaneously captures both low-frequency and high-frequency coefficients at the lowest scale. The remaining decoding layers and the incremental module jointly refine the high-frequency coefficients, leading to a novel and explainable registration framework. We summarize our contributions as:
\begin{itemize}
   \item  We embed a differentiable DWT layer before the convolutional encoder, which empowers our network to operate on the low-resolution representation of images in various frequency bands. 
    
    \item  We introduce an incremental module that features three levels of IDWT for coarse-to-fine deformation estimation.  This process takes advantage of both multi-scale prediction and DWT proprieties, avoiding the conventional multiple compositions and enhancing flexibility and interpretability. 
    \item Extensive results on 3D brain and cardiac registration show that our WiNet can achieve comparable accuracy with state-of-the-art pyramid methods such as LapIRN \cite{mok2020large}, PRNet++ \cite{kang2022dual}, and ModeT \cite{modelT} while using only 31.9\%, 25.6\%, and 23.2\% of their memory footprints (see Fig. \ref{fig:effici}), respectively.
     

\end{itemize}

\section{Methodology}
As illustrated in Figure \ref{fig:model}, WiNet takes a moving  $I_m$ and fixed  $I_f$ pair to predict the full-resolution deformation $\boldsymbol{\phi}$. The network comprises two parts: 1) in the encoder,  the DWT layer first decomposes each 3D input image into \textit{eight} components at various frequency sub-bands using the defined orthogonal wavelet filter, then the convolutional layers learn hierarchical feature representations from the decomposed components; 2) in the decoder,  the convolutional layers learn deformation related features at three scales. The incremental learning module takes the learned features from each decoding layer, progressively estimates scale-wise wavelet coefficients, and composes them into $\boldsymbol{\phi}$.


\subsection{DWT embedded Encoder}
Our encoder includes a parameter-free differentiable DWT layer and four convolutional contracting layers. The DWT layer decomposes an image $\mathbf{I}\in \mathbb{R}^{D \times H\times W}$ by applying the low-pass $\mathbf{L}$ and high-pass  $\mathbf{H}$  filters of an orthogonal Wavelet transform along the $H$, $W$, and $D$ dimensions, respectively. Therefore, $\mathbf{I}$ can be decomposed into one low-frequency component ($\mathbf{I}_{l l l}$) and seven high-frequency components ($\mathbf{I}_{llh}, \mathbf{I}_{lhl}, \mathbf{I}_{lhh}, \mathbf{I}_{hll}, \mathbf{I}_{hlh},  \mathbf{I}_{hhl}, \mathbf{I}_{hhh}$):


\begin{align}\label{eq_ref1}
  &\mathbf{I}_{l l l}   =\mathbf{L}  \obot \mathbf{L} \oright \mathbf{L}  \oleft  \mathbf{I},  \qquad
 \mathbf{I}_{l l h}   =\mathbf{L}  \obot \mathbf{L} \oright \mathbf{H}  \oleft  \mathbf{I},  \quad
 \mathbf{I}_{l h l}    =\mathbf{L}  \obot \mathbf{H} \oright \mathbf{L}  \oleft  \mathbf{I},  \quad  \nonumber \\
  &\mathbf{I}_{l h h}    =\mathbf{L}  \obot \mathbf{H} \oright \mathbf{H}  \oleft  \mathbf{I},  \quad
 \mathbf{I}_{h l l}   =\mathbf{H}  \obot \mathbf{L} \oright \mathbf{L}  \oleft  \mathbf{I},  \\
&  \mathbf{I}_{h l h}    =\mathbf{H}  \obot \mathbf{L} \oright \mathbf{H}  \oleft  \mathbf{I}, \quad
\mathbf{I}_{h h l}   =\mathbf{H}  \obot \mathbf{H} \oright \mathbf{L}  \oleft  \mathbf{I}, \quad
 \mathbf{I}_{h h h}   =\mathbf{H}  \obot \mathbf{H} \oright \mathbf{H}  \oleft  \mathbf{I} \quad \nonumber
\end{align}
where $\oleft$, $\oright$ and $\obot$ denote matrix multiplication in $H$, $W$, and $D$ dimensions, respectively; $\mathbf{L}$ and $\mathbf{H}$ are finite filters of wavelets that have truncated half-sizes for the dimension under filtering, e.g., $\left\lfloor\frac{H}{2}\right\rfloor \times H$ for the $H$ dimension.
\begin{align}\label{eq:lowpass_and_h}
\mathbf{L}  =\left(\begin{array}{ccccccc}
\cdots & \cdots & \cdots & & & & \\
\cdots & l_{-1} & l_0 & l_1 & \cdots & & \\
& & \cdots & l_{-1} & l_0 & l_1 & \cdots \\
& & & & & \cdots & \cdots
\end{array}\right),  \ 
\mathbf{H} =\left(\begin{array}{ccccccc}
\cdots & \cdots & \cdots & & & & \\
\cdots & h_{-1} & h_0 & h_1 & \cdots & & \\
& & \cdots & h_{-1} & h_0 & h_1 & \cdots \\
& & & & & \cdots & \cdots
\end{array}\right) ,
\end{align}
the values of $\mathbf{L}$ and $\mathbf{H}$ may vary based on the different orthogonal wavelet transform \cite{stollnitz1995wavelets}, while in this work, we only experiment with Haar and Daubechies.

As shown the gray dashed box in Fig. \ref{fig:model}, the DWT layer decomposes each of the full-resolution ${I}_m$ and ${I}_f$ into \textit{eight} frequency components, we then concatenate them together to $[16, \frac{D}{2}, \frac{H}{2}, \frac{W}{2}]$ as the input of the following convolutional layers. 
The resolution of feature maps in each contracting convolution layer is $\left[C, \frac{D}{2}, \frac{H}{2}, \frac{W}{2}\right]$, $\left[C, \frac{D}{4}, \frac{H}{4}, \frac{W}{4}\right]$, $\left[2C, \frac{D}{8}, \frac{H}{8}, \frac{W}{8}\right]$, and $\left[2C, \frac{D}{16}, \frac{H}{16}, \frac{W}{16}\right]$, respectively.

The benefits of the embedded DWT layer in our encoder are twofold. First, the decomposed various frequency sub-bands potentially improve the network's ability to capture both fine and coarse details.  Second, it naturally halves the spatial resolution of the $I_m$ and $I_f$ without sacrificing information, saving computational costs for subsequent convolutional layers.
We highlight that the DWT layer is not a pre-processing step, we made it a differentiable layer that allows back-propagation of gradients, and so is the IDWT layer.

\begin{figure*}[t]
    \centering
    \includegraphics[width=0.99\textwidth]{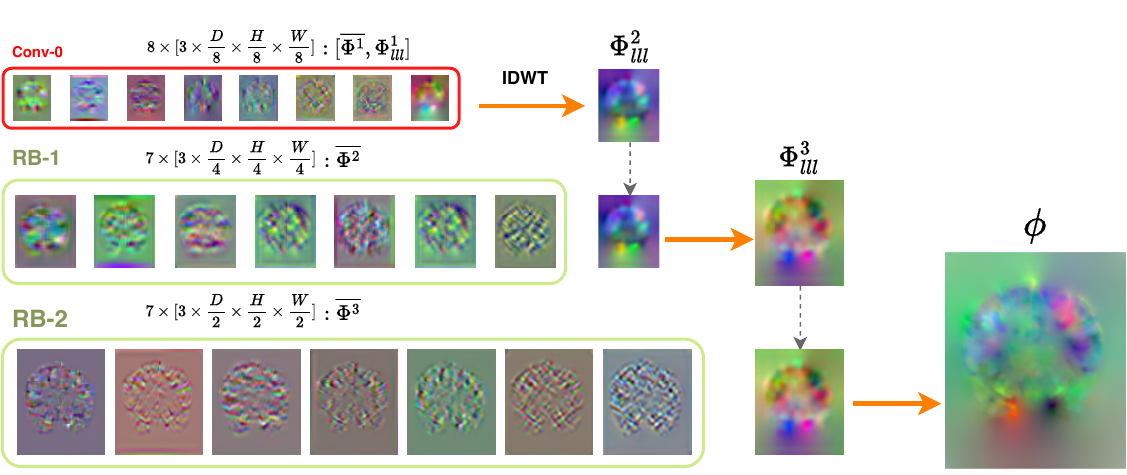}
    \caption{Illustration of the coarse-to-fine  incremental deformation learning module. }
    \label{fig:fig2}
\end{figure*}
\subsection{Incremental Deformation Learning Module \& Decoder}
The decoder consists of three expansive convolutional layers and an incremental deformation learning module. First, the convolutional decoding layers learn three different feature maps $F_1, F_2$ and $F_3$ with three different scales $\left[2C, \frac{D}{8}, \frac{H}{8}, \frac{W}{8}\right]$, $\left[2C, \frac{D}{4}, \frac{H}{4}, \frac{W}{4}\right]$ and $\left[C, \frac{D}{2}, \frac{H}{2}, \frac{W}{2}\right]$, respectively.  The incremental learning module will learn for each scale \textit{eight} different frequency sub-bands wavelet coefficients, which are denoted as $\Phi^{i}_{lll}, \Phi^i_{llh},\Phi^i_{lhl},\Phi^i_{lhh},\Phi^i_{hll},\Phi^i_{hlh}, \Phi^i_{hhl},\Phi^i_{hhh}$, where $i = 1, 2, 3 $ represents the $\frac{1}{8}$, $\frac{1}{4}$ and $\frac{1}{2}$ scale, respectively. Specifically, $\Phi^{i=1}_{lll}$ is the low-frequency coefficient, while the rest seven coefficients, denoted as $\overline{\Phi^i}$, contain high-frequency details.

As depicted in Figs.~\ref{fig:model} and ~\ref{fig:fig2}, the Conv-0 in our incremental module initially estimates \textit{eight} coefficients in the resolution of $\left[3, \frac{D}{8}, \frac{H}{8}, \frac{W}{8}\right]$ from feature $F_1$ to capture a coarse deformation. Then, the low-frequency $\Phi^{2}_{lll}$ is produced via the IDWT layer: $\Phi^{2}_{lll} = \text{IDWT}(\Phi^1) =\text{IDWT}([\Phi_{lll}^1,\overline{\Phi^1}]) $, where the IDWT operation, according to Eqs.~\eqref{eq_ref1} and~\eqref{eq:lowpass_and_h}, can be denoted as:
\begin{align}
{\Phi_{lll}^2} 
&=  \mathbf{L} \oleft \mathbf{L}  \oright  (\mathbf{L}\obot \Phi^1_{l l l} +  \mathbf{H} \obot \Phi^1_{h l l})  + \mathbf{L} \oleft \mathbf{H} \oright(\mathbf{L} \obot \Phi^1_{l l h}  +  \mathbf{H} \obot \Phi^1_{h l h})  \nonumber\\
&+ \mathbf{H} \oleft \mathbf{L} \oright (\mathbf{L} \obot \Phi^1_{l h l}  +  \mathbf{H} \obot \Phi^1_{hhl})  + \mathbf{H} \oleft \mathbf{H}  \oright ( \mathbf{L} \obot \Phi^1_{l h h} + \mathbf{H} \obot \Phi^1_{h h h}) .
\end{align}
Similarly, \( \Phi^3_{lll} \) can be produced by performing IDWT with \(\Phi^{2}_{lll}\) and the rest \textit{seven} $[3, \frac{D}{4}, \frac{H}{4}, \frac{W}{4}]$ high-frequency coefficients, i.e, $\Phi^{3}_{lll} =\text{IDWT}([\Phi_{lll}^2,\overline{\Phi^2}])$, then the final ${\phi}$ can be obtained as ${\phi} = \text{IDWT}([\Phi_{lll}^3, \overline{\Phi^3}])$.The primary challenge here is how can we learn the seven high-frequency $\overline{\Phi^2}$ and $\overline{\Phi^3}$ using $\overline{\Phi^1}$?

To address the challenge, we propose two refinement blocks (RB-1 and 2). As in Fig.~\ref{fig:model}, RB-1 incrementally refines $\overline{\Phi^1}$ to  $\overline{\Phi^2}$ and RB-2 refines $\overline{\Phi^2}$ to $\overline{\Phi^3}$. Specifically, 1) the up-sampling operation in RB-1 (the yellow trapezoid) first converts the seven $\left[3, \frac{D}{8}, \frac{H}{8}, \frac{W}{8}\right]$ coefficients of $\overline{\Phi^1}$ into the resolution of $\left[3, \frac{D}{4}, \frac{H}{4}, \frac{W}{4}\right]$; 2) the conv layer (red arrow) takes learned feature $F_2$ and estimates also seven $\left[3, \frac{D}{4}, \frac{H}{4}, \frac{W}{4}\right]$ coefficients, which serve as the complementary residuals to the upsampled coefficients; 3) then both the upsampled and learned coefficients are concatenated together in the frequency order; 4) and a group convolution (group=7, illustrated as the red dashed arrows) is used to output the refined $\overline{\Phi^2}$. The process can be denoted as:
\begin{align}
\overline{\Phi^2} &= GC(Cat[Conv({F}_2), UP(\overline{\Phi^1})])
\end{align}
where $UP$, $Conv$, $Cat$, and $GC$ denote upsampling, convolution, concatenation by frequency order, and group convolution, respectively. Likewise, we estimate $\overline{\Phi^3}$ from $\overline{\Phi^2}$ and $F_3$ with RB-2. Following the same procedure, the full-resolution deformation is produced by $\boldsymbol{\phi} = \text{IDWT}([\Phi_{lll}^3, \overline{\Phi^3}])$.

\subsection{Loss Functions}
For diffeomorphic registration (WiNet-Diff), the output of WiNet-Diff is the stationary velocity fields $\mathbf{v}$ and we employ seven scaling and squaring layers \cite{DARTEL} to integrate the deformation field  $\boldsymbol{\phi} = \exp(\mathbf{v})$. The training loss of our network is 
$\mathcal{L}(\boldsymbol{\theta})= \mathcal{L}_{Sim}(I_m \circ(\boldsymbol{\phi}(\boldsymbol \theta)+\mathrm{Id}), I_f) +\lambda\left\|\nabla \boldsymbol{\phi}(\boldsymbol \theta)\right\|_2^2$ or $\mathcal{L}(\boldsymbol{\theta})= \mathcal{L}_{Sim}(I_m \circ \exp(\mathbf{v}(\boldsymbol \theta)),I_f) +\lambda\left\|\nabla \mathbf{v}(\boldsymbol \theta)\right\|_2^2$, where $\boldsymbol\theta$ represents the learnable network parameters, $\circ$ refers the warping operator, $\mathrm{Id}$ is the identity grid, $\nabla$ is the first order gradient and $\lambda$ is the weight of the regularization term.The term $\mathcal{L}_{\text{sim}}(\cdot)$ quantifies the similarity between the warped moving image $I_m \circ \boldsymbol{\phi}$ and the fixed image $I_f$. The second term are the smoothness regularization on $\boldsymbol{\phi}$.
\section{Experiments}
\noindent \textbf{Datasets.} One 3D brain MRI dataset (IXI) and  one 3D cardiac MRI dataset (3D-CMR) were utilized. The pre-processed IXI\footnote{\href{https://brain-development.org/ixi-dataset/} {https://brain-development.org/ixi-dataset/}} dataset \cite{chen2022transmorph} comprises 576 MRI scans (160×192×224) from healthy subjects. We followed the same protocol with partitions 403, 58, and 115 for the training, validation, and testing sets, respectively. Atlas-based brain registration was conducted for each scan using a generated atlas from \cite{kim2021cyclemorph}. The 3D-CMR  \cite{thorley2021nesterov,jia2021learning,duan2019automatic} contains 220 pairs of End-diastolic (ED) and End-systolic (ES) scans. All MRI scans were resampled to the same resolution $1.2 \times 1.2 \times 1.2\, \text{mm}^3$ and center-cropped to dimensions of 128×128×96. We randomly split it into 100 training, 20 validation, and 100 testing sets. \\


\begin{figure*}[ht]
    \centering
    \includegraphics[width=0.99\textwidth]{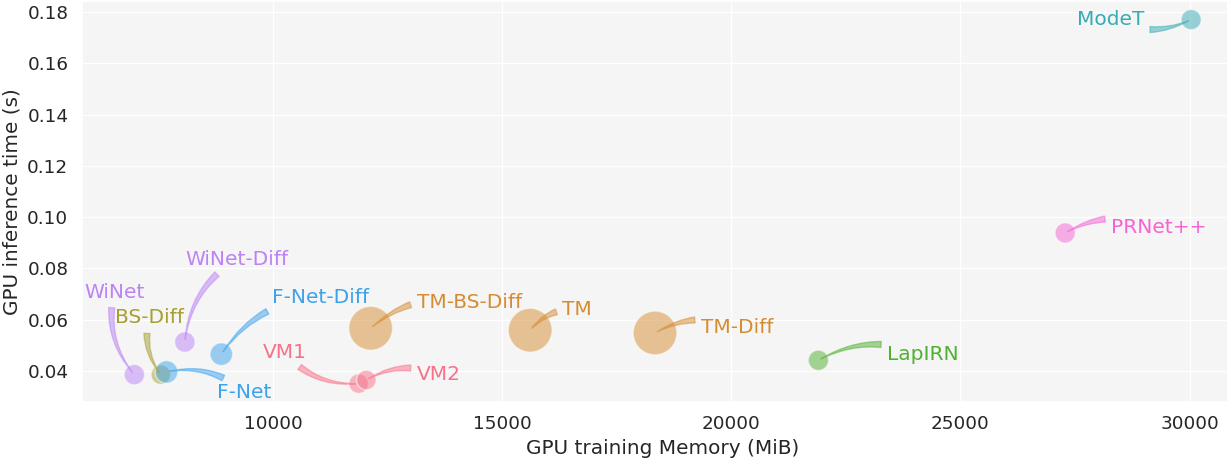}
    \caption{Comparison of the number of parameters (the area of circles), GPU training memory on IXI, and averaged GPU inference time on the same machine. Fourier-Net, TM-B-Spline-Diff, and B-Spline-Diff are abbreviated as F-Net, TM-BS-Diff, and BS-Diff, respectively. Our WiNet requires the lowest memory (6977 MiB) for training.}
    \label{fig:effici}
\end{figure*}
\begin{figure*}[t]
    \centering\includegraphics[width=0.995\textwidth]{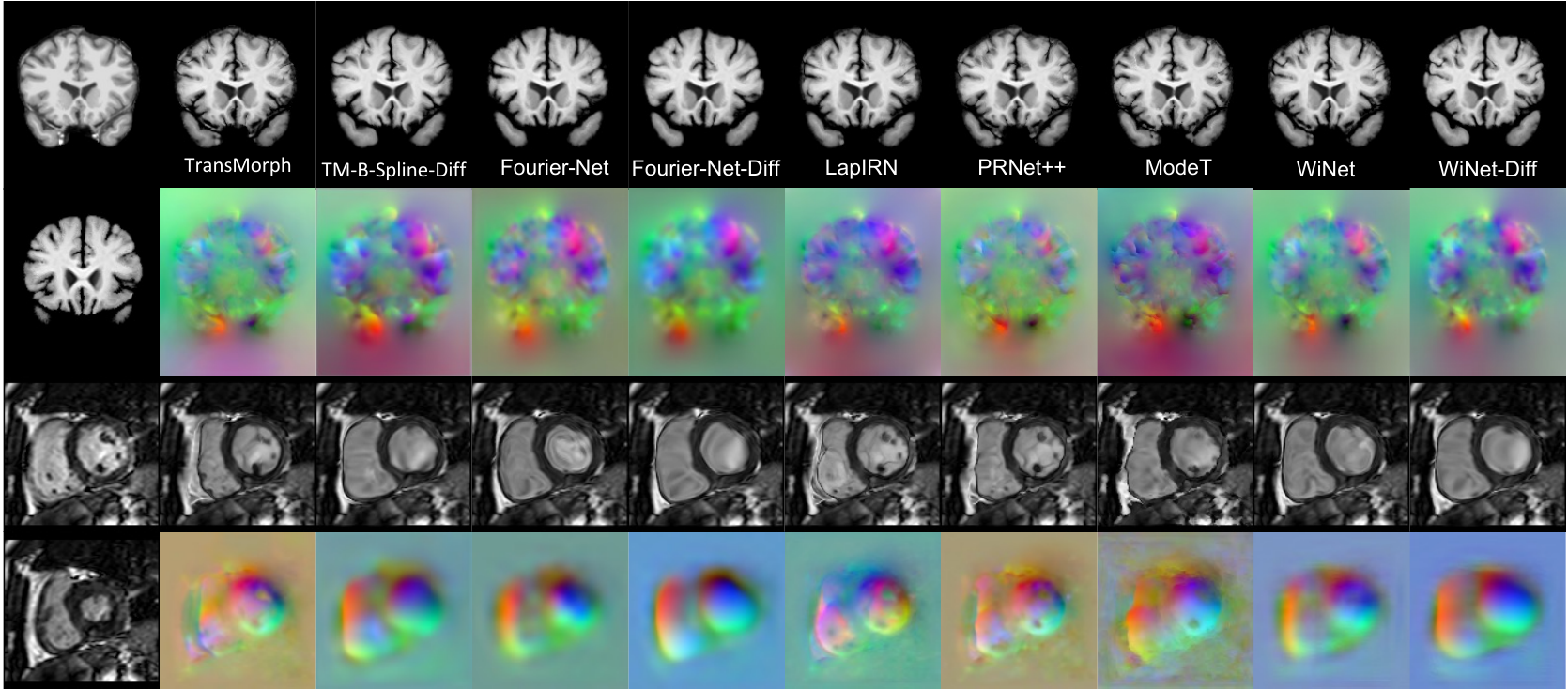}
    \caption{Visual comparisons on  IXI and 3D-CMR. Column 1: Fixed image, Moving image. Columns 2-10: warped moving images, displacement fields as RGB images. }
     \label{fig:visual}
\end{figure*}
\noindent \textbf{Evaluation Metrics.} We use the Dice score, Hausdorff distance (HD), and the percentage of negative values of the Jacobian determinant ($|J|_{ < 0}\%$) to evaluate the registration accuracy. We use the number of parameters, GPU training memory in Mebibytes (MiB), and the average GPU inference time in seconds (s) to measure model efficiency. \\
\begin{table*}[!h]
\centering
    \caption{Registration performance comparison on the IXI and 3D-CMR datasets.}
\begin{small}
        \begin{tabular}{lcccccc}
        \toprule[1pt]
        & \multicolumn{2}{c}{\textbf{IXI}} & \multicolumn{3}{c}{\textbf{3D-CMR}} \\
      \cmidrule(lr){2-3} \cmidrule(lr){4-6}
        Method & Dice$\uparrow$ & $|J|_{ < 0}\%$  & Dice$\uparrow$ & HD$\downarrow$ & $|J|_{ < 0}\%$ \\
        \midrule
        Affine   & .386$\pm$.195 &-  &  .493$\pm$.043  & 8.40$\pm$0.89& -\\
    SyN \cite{avants_ANTS} & .645$\pm$.152 & \textless{}0.0001  & .721$\pm$ .051 & 6.98$\pm$1.23 & 1.388$\pm$0.467  \\
    \hline
    VM-1 \cite{balakrishnan2019voxelmorph}  & .728$\pm$.129 & 1.590$\pm$0.339 & .723$\pm$.032 & 6.53$\pm$1.00 & 0.783$\pm$0.325  \\
    VM-2 \cite{balakrishnan2019voxelmorph}  & .732$\pm$.123 & 1.522$\pm$0.336 & .723$\pm$.032 & 6.71$\pm$1.04 & 0.695$\pm$0.255 \\
    TM \cite{chen2022transmorph} & .754$\pm$.124 & 1.579$\pm$0.328 & .732$\pm$.029 & 6.51$\pm$0.99 & 0.811$\pm$0.231 \\
    TM-Diff \cite{chen2022transmorph}  & .594$\pm$.163 & \textless{}0.0001  &.630$\pm$.029 &8.15$\pm$0.80 & \textbf{0.0$\pm$0.0} \\
    TM-B-Spline-Diff \cite{chen2022transmorph} & .761$\pm$.122 & \textless{}0.0001 & .810$\pm$.026 & \textbf{5.84$\pm$0.73} & \textless{}0.0001 \\ 
    \hline
    B-Spline-Diff \cite{qiu2021learning} & .742$\pm$.128 & \textless{}0.0001  &  .809$\pm$.033 & 6.04$\pm$0.73 & \textbf{0.0$\pm$0.0}\\
    Fourier-Net \cite{jia2023fourier} & .763$\pm$.129 & 0.024$\pm$0.019 &.814$\pm$.024 & 6.00$\pm$0.77 & 2.137$\pm$0.658  \\
    Fourier-Net-Diff \cite{jia2023fourier} &.761$\pm$.131 & \textbf{0.0$\pm$0.0}   & .827$\pm$.027 & 6.08$\pm$0.85 & \textless{}0.0003  \\\hline
    LapIRN \cite{mok2020large}& .760$\pm$.127&0.312$\pm$0.106 &.760$\pm$.028 &6.61$\pm$0.90 &1.100$\pm$0.299 \\
    PRNet++ \cite{kang2022dual} &.755$\pm$.130&1.052$\pm$0.302& .749$\pm$.031 &6.74$\pm$1.02 &0.257$\pm$0.115\\
    ModeT \cite{modelT}  & .758$\pm$.125 & 0.114$\pm$0.057 & .774$\pm$.030& 6.72$\pm$0.96 &0.041$\pm$0.017  \\\hline
    WiNet & .763$\pm$.125 &  0.874$\pm$0.284  & .814$\pm$.027 & 6.06$\pm$0.66 & 2.160$\pm$0.612 \\
    WiNet-Diff &\textbf{.765$\pm$.125}&\textbf{0.0$\pm$0.0}& \textbf{.832$\pm$.023} &{5.86$\pm$0.68}  & 0.007$\pm$0.007\\
            \bottomrule[1pt]
        \end{tabular}
\end{small}
    \label{tab:table1}
\end{table*}

\noindent\textbf{Implementation Details.} Our method is implemented using PyTorch. Both the training and testing phases are deployed on an A100 GPU with 40GB VRAM. All models are optimized using Adam with a learning rate of \(1 \times 10^{-4}\) and batch size of 1 for  1000 epochs. On IXI, we employed the Normalized Cross-Correlation (NCC) loss with \(\lambda =2 \). For 3D-CMR, we employed the Mean Squared Error (MSE) loss with \(\lambda =0.01 \). We used $C=32$ for all experiments. \textcolor{black}{We used Haar and Daubechies wavelet transform for the IXI and 3D-CMR datasets.}\\

\noindent \textbf{Comparative Methods.} We compare our method with a series of state-of-the-art, including conventional SyN \cite{pmid17659998}, U-Net-based methods (VoxelMorph (VM) \cite{balakrishnan2019voxelmorph}, TransMorph (TM) \cite{chen2022transmorph}),  model-driven methods (B-Spline-Diff \cite{qiu2021learning}, Fourier-Net \cite{jia2023fourier}), and three pyramid methods (LapIRN \cite{mok2020large}, PR++ \cite{kang2022dual},  ModeT \cite{modelT}). All methods are trained with the official released code and the optimal parameters are tuned on the validation sets.\\

\noindent \textbf{Quantitative and Qualitative Analysis.} Table \ref{tab:table1} shows the numerical results of compared methods on the IXI and 3D-CMR datasets. It can be seen that our method outperforms all compared methods in terms of Dice. WiNet outperforms the iterative SyN with margins of 12\% and 11\% Dice on IXI and 3D-CMR, respectively. WiNet surpasses model-driven approaches, achieving improvements of $0.2\%$ (Fourier-Net) and $0.5\%$ (Fourier-Net-Diff) in Dice for the IXI and 3D-CMR and securing the second-best in terms of HD.
 Compared with pyramid methods, our method achieves the increase of Dice by at least $0.5\%$ and $5.8\%$ on IXI and 3D-CMR. Notably, Fig.~\ref{fig:effici} clearly illustrates that WiNet requires less GPU memory than pyramid methods in training, Our method exhibits efficiency, consuming only 31.9\%, 25.6\%, and 23.2\% of the memory compared to LapIRN, PRNet++, and ModeT, respectively, while presenting faster speed than all three of them. Moreover, our WiNet-Diff can achieve 0\% and 0.007\%  $|J|_{<0}\%$ on IXI and 3D-CMR, respectively, resulting in a plausible deformation field compared to our WiNet, as assessed in Fig. \ref{fig:visual}. Additionally, we can observe that the estimated deformation fields from model-driven methods including B-Spline-Diff, Fourier-Net, and our WiNet tend to be smoother than U-Net-based methods (TM) and pyramid methods.
\section{Conclusion}
In this paper, we propose a model-driven WiNet to perform coarse-to-fine 3D medical image registration by estimating various frequency wavelet coefficients at different scales. Inheriting the properties of the wavelet transform, WiNet learns the low-frequency coefficients explicitly and high-frequency coefficients incrementally, making it explainable, accurate, and GPU efficient. 
\begin{credits}
\subsubsection{\ackname} The research were performed using the Baskerville Tier 2 HPC service. Baskerville was funded by the EPSRC and UKRI through the World Class Labs scheme (EP/T022221/1) and the Digital Research Infrastructure programme (EP/W032244/1) and is operated by Advanced Research Computing at the University of Birmingham.

\subsubsection{\discintname}
The authors have no competing interests to declare that are
relevant to the content of this article. 
\end{credits}
\bibliographystyle{splncs04}
\bibliography{my}

\end{document}